\documentclass{IOS-Book-Article}     
\usepackage{mathptmx}
\usepackage[T1]{fontenc}
\usepackage{times}%
\usepackage{graphicx}

\usepackage{multirow}
\usepackage{amsmath}
\usepackage{amsfonts}
\usepackage{algorithmic}

\newcommand{\indep}{\perp \!\!\! \perp}

\begin{document}
\begin{frontmatter}          
%
\title{Unsupervised Pairwise Causal Discovery on Heterogeneous Data using Mutual Information Measures}
\runningtitle{}

\author[A]{\fnms{Alexandre} \snm{Trilla}},
\author[B]{\fnms{Nenad} \snm{Mijatovic}}
\runningauthor{}
\address[A]{Alstom, Santa Perp\`etua de la Mogoda, Barcelona, 08130, Spain}
\address[B]{Alstom, Saint Ouen, Paris, 93482, France}
\begin{abstract}
A fundamental task in science is to determine the underlying
causal relations because it is the knowledge of this functional
structure what leads to the correct interpretation of an effect 
given the apparent associations in the observed data.
In this sense, 
Causal Discovery is a technique that tackles this
challenge by analyzing the statistical properties of the 
constituent variables.
In this work, we target the generalizability of the 
discovery method by following a reductionist approach that 
only involves two variables, i.e., the pairwise or bi-variate
setting.
We question the current (possibly misleading)
baseline results on the basis that they were obtained through 
supervised learning, which is arguably contrary
to this genuinely exploratory endeavor. In consequence, we
approach this problem in an unsupervised way, using robust Mutual
Information measures, and observing the impact of the different variable
types, which is oftentimes ignored in the design of solutions.
Thus, we provide a novel set of standard unbiased results
that can serve as a reference to guide future discovery tasks 
in completely unknown environments.
\end{abstract}

\begin{keyword}
Causal Discovery, Functional Causal Model, Unconditional Independence Test, Mutual Information, Unsupervised Learning, Heterogeneous Data
\end{keyword}

\end{frontmatter}


\section{Introduction}
Causal Discovery methods are able to identify the causal structure
from the joint distribution of the data by introducing assumptions 
that restrict the model of their generating process~\cite{Pearl09}.
In a multivariate setting, the traditional constraint-based and 
score-based methods exploit conditional independence 
relationships in the data~\cite{Glymour19}.
These approaches have found a great deal of success in 
challenging environments such as biology and the Earth
sciences~\cite{Runge19Nature}. Nevertheless,
they do not necessarily provide complete causal information 
because they output Markov Equivalence Classes, i.e., a set of
causal structures that satisfy the same conditional independence
statements. Moreover, they cannot handle isolated cause-effect
settings that lack the diversity of other variables for the 
conditional tests. Therefore, the single reduced cause-effect association,
also known as the pairwise or bivariate scenario, constitutes an 
essential building block of more complex causal structures. 

To distinguish cause from effect in this pairwise setting, 
one needs to find a way to capture the asymmetry between the
two variables~\cite{Glymour19}. In this sense,
computational methods based on properly defined Functional Causal 
Models (FCMs) are able to distinguish different structures in the 
same equivalence class.
A FCM represents the effect variable $Y$ as a function of its direct
causes $X$ and some noise term $\varepsilon$, i.e., 
$Y = f(X, \varepsilon)$, where $\varepsilon$ is
independent of $X$. Thanks to the restricted functional classes,
the causal direction between $X$ and $Y$ is generally identifiable because
the independence between the noise and the cause holds
only for the true causal direction and is violated for the wrong
direction~\cite{Peters17} (unless the variables are jointly Gaussian,
which render the orientation unidentifiable). Additionally, the
identifiability of pairwise FCM generalizes to the identifiability 
of multivariate FCM~\cite{Peters2011b}.

In this work, we argue that some reference benchmark results for pairwise 
causal discovery could be misleading because they tackle this genuinely 
exploratory objective as a classical supervised classification problem, 
inducing them to possibly learn specific contextual details beyond causality 
that tend to overestimate their actual performance~\cite{Guyon13}.
Moreover, there exists a tight relationship between the nature of data 
and the specific tools from statistics that could also accrue the estimation 
of skewed results, which was not originally addressed. Consequently, we 
propose a reevaluation of the benchmark driven by the heterogeneity 
of the data and also by following an unsupervised 
learning approach. Our method leverages linear functional causal models and 
combines different unconditional independence tests based on robust
Mutual Information measures.
The paper is organized as follows: Section~\ref{secBackground}
introduces the fundamental concepts for pairwise causal
discovery, Section~\ref{secMethod} describes the enhanced method
proposed in this work, Section~\ref{secResults} shows the novel
benchmarked results, Section~\ref{secDisc}
discusses the approach from a causal perspective, along with its
limitations, and
Section~\ref{secConc} concludes the article.

\section{Background} 
\label{secBackground}
This section reviews the fundamental concepts that support the
discovery of causality in the bivariate or pairwise setting.

\subsection{Identifiability of a Linear Model with Non-Gaussian Noise}
The fundamental principle that enables the identification of cause
and effect lies in the asymmetry of the association between them: 
the causal direction, i.e., from cause to effect, is functionally 
simpler than the anticausal one, i.e., from effect to cause, which
requires a more complex application~\cite{Glymour19}.
Thus, a given model with limited capacity will be able to 
faithfully capture the expressiveness 
of \emph{only} one of these two causal orientations.

Linear additive noise models assume that the effect $Y$ is a linear 
function of the cause $X$ plus a noise term $\varepsilon$ that
is independent of the cause, formally defined as: $Y = bX + \varepsilon$.
The linear FCM is learnable if at most one of $X$ and $\varepsilon$
is Gaussian~\cite{Shimizu06}. Pairwise discovery approaches such as this one are regarded to be very flexible for identifying causal models in the general case because they don't require a third variable for conditioning.
In specific scenarios such as multivariate time series, they can even
improve the performance of traditional constraint-based
approaches such as the Peter-Clark algorithm and others that
utilize Granger causality~\cite{Runge19Nature}.

The linear model with non-Gaussian noise can be estimated from 
(unconfounded) observational data by exploiting the inherent 
asymmetry between cause and effect through a regression as 
follows: for both directions (i.e., causal and anticausal), the 
FCM is fit; then, the amount of association between the estimated 
noise term (i.e., the regression residual) and the hypothetical 
cause is computed; finally, the direction assignment which gives an 
independent noise term is considered plausible~\cite{Glymour19}.

Obviously, the cornerstone of this regression-based method is the
measure of dependence between the regression residual and its
associated potential cause. The next section 
describes a solution to estimate the strength of this association.

\subsection{Mutual Information as a Measure of Association}
The amount of association between the regression residual $RR$ and
the hypothetical cause $HC$ signals the causal direction. However,
these variables are uncorrelated by construction of the regression. To 
properly quantify this dependence, the Mutual Information (MI) measure 
provides a reliable indicator, which is defined as

\begin{equation}
	MI(RR,HC) = \sum_{RR,HC} P_{RR,HC} \log \left ( \frac{P_{RR,HC}}{P_{RR} \, P_{HC}} \right ) \ ,
\label{eqMI}
\end{equation}

where $P_{RR,HC}$ represents the (discrete) joint probability between
$RR$ and $HC$, and $P_{RR}$ and $P_{HC}$ represent their marginal
distributions, respectively.

For illustrative purposes, the following variable dependencies 
are defined. First, a random Uniform noise sample $U$ is
independently assigned to $Z$, $X_{ind}$ and $Y_{ind}$. Then,
the following structural variable associations are created: 
$Y_{dep} \leftarrow X_{ind} + Z$, $X_{conf} \leftarrow X_{ind} + Z$,
and $Y_{conf} \leftarrow Y_{ind} + Z$, which yield the subsequent relationships:
\begin{description}
\item[Causal] Between $X_{ind}$ and $Y_{dep}$: $X \rightarrow Y$
\item[Anticausal] Between $Y_{dep}$ and $X_{ind}$: $Y \leftarrow X$
\item[Independent] Between $X_{ind}$ and $Y_{ind}$: $X \indep Y$
\item[Confounded] Between $X_{conf}$ and $Y_{conf}$: $X \leftrightarrow Y$
\end{description}

\begin{figure}[htb]
\centering
\includegraphics[width=0.5\textwidth]{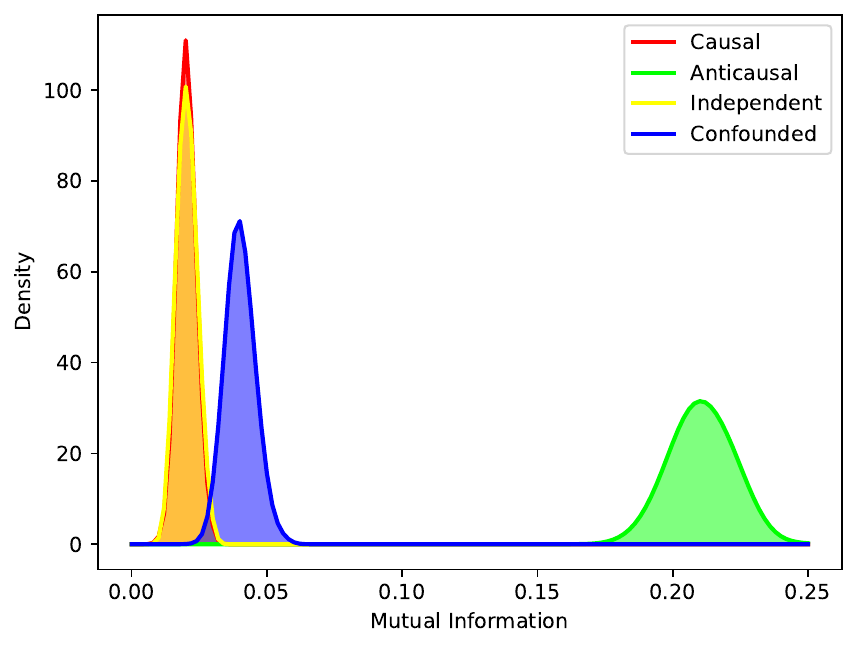}
	\caption{Distribution of Mutual Information between
	the regression residual and the related hypothetical cause
	for the different causal structures. The low-value
	overlap (i.e., MI$<$0.025) between the
	Causal and the Independent structures illustrates the
	independence between the noise and the cause.
	The density of the real-valued MI indicator has been estimated
	with a smoothed Gaussian kernel.}
\label{figMIDist}
\end{figure}

Figure~\ref{figMIDist} shows the distribution of MI between
the regression residual and the hypothetical cause considering
the preceding types of structural variable associations for a random
sample of 1000 instances. The different density plots
show how the causal orientation can be reliably recovered using
MI when the model assumptions are respected. Note that
the overlap between Causal and Independent structures 
will be disambiguated when both experiments are conducted regarding 
the different causal direction hypotheses.
Finally, MI is also able to detect latent confounding, which is 
oftentimes assumed (and required) not to occur for 
discovery~\cite{Glymour19}, i.e., the causal 
sufficiency principle~\cite{vanDiepen23}. This feature 
adds value to the robustness of the MI indicator.

While the distribution of MI is a useful tool to visualize the
different causal structures, there is still a missing criterion to 
make a classification decision in order to construct the underlying
causal structure from the data. The next section addresses this point.

\subsection{Pearson's $\chi^2$ as an Unconditional Independence Test}
\label{subPearsonsChi2}
In a real-word scenario, one cannot expect to get a null measure
of association because of sampling noise, the sample size, etc.
In this case, the use of an (unconditional) independence
test is the proper way to handle this situation. In general,
this is known as the REgression with Subsequent Independence Test
(RESIT) procedure~\cite{Peters14RESIT}. 
Once the independence test is in place, a confidence interval needs 
to be observed to obtain actionable results. To this end, Pearson's 
$\chi^2$ test is here introduced as a fundamental 
method to inform the decision process~\cite{Pearson1900}.

The Pearson's $\chi^2$ test is a statistical hypothesis test used 
in the analysis of contingency tables for categorical variables.
It is used here to determine whether there is a statistically 
significant difference between the expected frequencies $E$
for independence (i.e., the marginal probabilities)
and the observed frequencies $O$ in one or more categories $i$
of the contingency table $(RR,HC)$, formally defined as
\begin{equation}
	\chi^2(RR,HC) = \sum_{i \in (RR,HC)} \frac{(O_i - E_i)^2}{E_i} \ .
\label{eqChi2}
\end{equation}

The test is valid as long as its statistic is $\chi^2$ distributed 
under the null hypothesis.
Note that $\chi^2(RR,HC)$ is in its essence related to the MI measure of 
association as is used for the discovery of causal structure.

While the $\chi^2$ test is most suitable for discrete variables~\cite{Reshef16},
a general
causal discovery method should be able to seamlessly deal with 
continuous variables. Typically, a uniform grid is applied to discretize
a real-valued space prior to conducting the test.
Following the former illustrative example, Table \ref{tabChi2}
shows the p-values for the different causal structures.
Once a threshold is introduced through the confidence interval,
e.g., p$<$0.05 for a given number of degrees of freedom related to
the resolution of the grid,
passing the test becomes the rule for making further decisions.
Note that in all cases, there is a unique combination of results
for the two hypothesized causal directions that identifies the true
underlying causal structure.
Also note that the threshold may be adjusted to require further decision strength because the closer the variables get to a Gaussian distribution, the harder it is to distinguish the direction of causation~\cite{Glymour19}.

\begin{table}[h] \small  
	\begin{center}  
	\caption{Pearson's $\chi^2$ p-values for the independence tests
	between the regression residuals and the hypothesized causal directions
	for the different illustrative structures, considering 10 bins in
	the uniform grid.\\}
	\label{tabChi2}
	\begin{tabular}{ l | c c c c}
		\hline
		\multirow{2}{1.65cm}{\textbf{Hypothesis Direction}} & \multicolumn{4}{c}{\textbf{True Structure}}\\
		\cline{2-5}
		& \textbf{Causal} & \textbf{Anticausal} & \textbf{Independent} & \textbf{Confounded}\\ 
		\hline
		Causal & 0.5027 & 0.0000 & 0.4847 & 0.0025\\
		Anticausal & 0.0000 & 0.5027 & 0.4639 & 0.0025\\ \hline
	\end{tabular}
	\end{center}
\end{table}

While these results
may be sufficient as an indication, they may also have some 
limitations for discovering causal associations between numerical variables.
The next section addresses this shortcoming and describes
an enhanced solution for this type of variables.

\subsection{Total Information Coefficient as a Robust Independence Test}
\label{subTIC}
The Total Information Coefficient (TIC) is a robust independence
test based on MI for real-valued variables 
that features the two important heuristic properties of generality and 
equitability~\cite{Reshef11}, which are defined as follows:

\begin{description}
\item[Generality] With sufficient sample size the statistic 
should capture a wide range of interesting associations, not limited 
to specific function types. 
\item[Equitability] The statistic should give similar scores to 
equally noisy relationships of different types.
\end{description}

\begin{figure}[htb]
\centering
\includegraphics[width=0.6\textwidth,trim=0cm 21cm 0cm 2.5cm, clip]{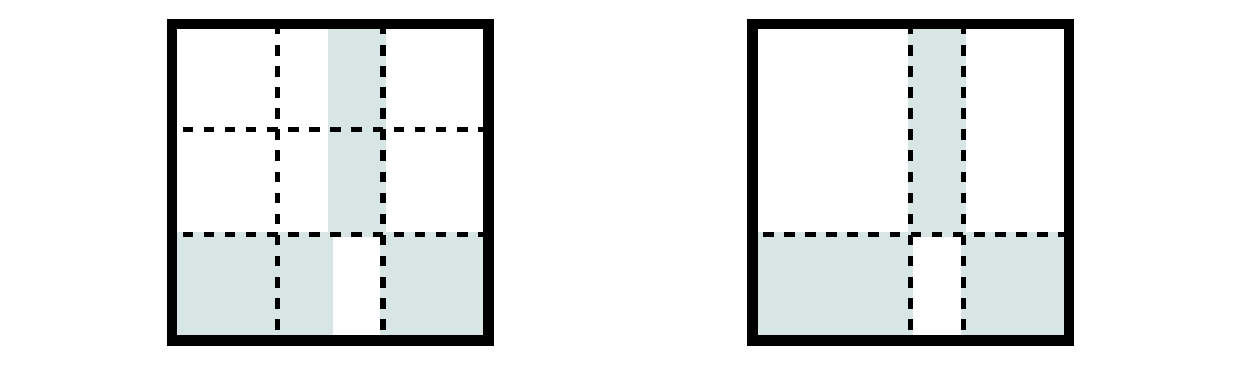}
	\caption{A schematic illustrating the difference between 
	the gridding strategies (shown in dotted lines)
	introduced by the independence tests. 
	(Left) Uniform 3-by-3 grid as is used by the $\chi^2$ test. 
	(Right) Optimum 2-by-3 grid $G$ as is used by the TIC test.}
\label{figGrids}
\end{figure}

For a pair of $(RR,RC)$ variables, the TIC algorithm applies a search
procedure to partition their joint probability function and find the
grid with the highest induced MI.
TIC operates by summing over optimal grids $G$ on the joint 
density distribution of the two variables, such that it exhibits 
a stronger power 
against independence~\cite{Reshef16}, formally expressed as
\begin{equation}
	TIC(RR,HC) = \sum_G \frac{MI((RR,HC)|_G)}{\log \parallel G \parallel} \ ,
\label{eqTIC}
\end{equation}
where $\parallel G \parallel$ denotes the minimum of the number 
of rows and columns of $G$. This can be viewed as a regularized 
version of MI that penalizes complicated grids.
Figure~\ref{figGrids} shows qualitatively the impact of the
optimum grids $G$.

The performance of TIC has been positively rated with
other successful approaches for causal discovery dealing with
continuous variables, such as the
Hilbert-Schmidt independence criterion~\cite{Gretton07}.
Finally, an additional motivation for
considering TIC in this work is to increase the odds of success
in case of model hypothesis violations, i.e., the functional form
of the relationship $Y = f(X)$, which is crucial for causal
identification.

\section{Method}
\label{secMethod}
In the pursuit of a general approach to causal discovery
combining several methods~\cite{vanDiepen23,Zeng22LinMixDat},
one common point of success is found in the use
of information-theoretic measures, such as Mutual Information,
to quantify the amount of regularity in the 
data~\cite{Giori22}. 
However, the nature of the variables, i.e., discrete or continuous, 
can be problematic in an heterogeneous context if the statistical 
methods are confused~\cite{Zeng22LinMixDat}.
Moreover, supervised learning approaches are excluded from the
general objective because these methods do not yet work as standalone 
techniques for causal learning~\cite{Peters17}. Therefore,
the desideratum in causal discovery is to have methods that work on a 
broad range of problems under different conditions with relaxed
assumptions~\cite{Duong22}, ideally showing a 
certain degree of robustness regarding violations of the model 
hypothesis~\cite{Montagna23}.

In this work we propose a general method to discover the causal 
structure in heterogeneous data based on a bivariate linear FCM by
implementing a flexible RESIT procedure, see 
Section~\ref{subPearsonsChi2}, where the
unconditional independence test, e.g., $\chi^2$ or TIC,
also see Section~\ref{subTIC},
is driven by the nature of the variables involved. The proposed 
strategy is described with comments in Algorithm~\ref{algoRESFIT}.
Note that the orientation decision rules follow from the evidence 
given by the illustrative setting in Section~\ref{subPearsonsChi2}, 
which is regarded as self-evident.
Such explicitly-stated logic rules are meant to increase the 
interpretability and explainability of the proposed method.
Also note that discrete variables are assimilated to real-valued variables for computing the regression residuals on which the RESIT method is based. Finally,
since our contribution targets the problem of applying one single discovery technique on data with different types of variables, and we focus our effort at the integration level rather than at the fundamental level, we refer the interested reader to the references for the specific methods being integrated for further details about their comparison to other methods from the literature.
We rely on the advice from Peters, Reshef, and 
colleagues~\cite{Peters17,Reshef16}, to select the best methods that
we integrate in RESFIT in order to cover a broad range of techniques, 
and we provide an analysis on how these separate tools perform when
they are put together.

\begin{algorithm}
\caption{REgression with Subsequent Flexible Independence Test (RESFIT).
	The traditional RESIT procedure is here used as an internal
	function (underscore-prepended).\\}
\label{algoRESFIT}
\begin{algorithmic}
\STATE {\bf Input:} A, B {\tt //Potential causally-related variables}
\STATE {\bf Output:} Causal, Anticausal, Independent, Confounded {\tt //Causal structure}
\STATE
\STATE function {\bf RESFIT} (A, B):
\STATE \ \ \ \ histAB = Histogram2D(A, B) {\tt //Joint variable distribution}
\STATE \ \ \ \ varTypes = JointTypes(histAB) {\tt //Cat., bin., num., mix.}
\STATE \ \ \ \ uit = SelectUIT(varTypes) {\tt //Uncond. Indep. Test: chi2, tic}
\STATE \ \ \ \ pCH = \_\_RESIT(A, B, uit) {\tt //p-value of Causal hypothesis}
\STATE \ \ \ \ pACH = \_\_RESIT(B, A, uit) {\tt //p-value of Anticausal hypothesis}
\STATE \ \ \ \ ci = 0.05 {\tt //Confidence interval}
\STATE \ \ \ \ {\tt //---- Decision Rules ------------------------------------------}
\STATE \ \ \ \ {\bf if} {(pCH $>$ ci) {\bf and} (pACH $<$ ci)} {\bf then}
\STATE \ \ \ \ \ \ \ \ {\bf return} Causal {i\tt //Reject Anticausal independence}
\STATE \ \ \ \ {\bf else if} {(pCH $<$ ci) {\bf and} (pACH $>$ ci)} {\bf then}
\STATE \ \ \ \ \ \ \ \ {\bf return} Anticausal {\tt //Reject Causal independence}
\STATE \ \ \ \ {\bf else if} {(pCH $>$ ci) {\bf and} (pACH $>$ ci)} {\bf then}
\STATE \ \ \ \ \ \ \ \ {\bf return} Independent {\tt //Cannot reject any independence}
\STATE \ \ \ \ {\bf else} 
\STATE \ \ \ \ \ \ \ \ {\bf return} Confounded {\tt //Reject all independence}
\STATE
\STATE function {\bf \_\_RESIT} (var1, var2, UIT):
\STATE \ \ \ \ HC = var1 {\tt //Hypothetical Cause}
\STATE \ \ \ \ HE = var2 {\tt //Hypothetical Effect}
\STATE \ \ \ \ lr = LinearRegression(HE, HC) {\tt //Linear function restriction}
\STATE \ \ \ \ RR = HE - lr(HC) {\tt //Regression Residual}
\STATE \ \ \ \ {\bf return} UIT(RR, HC) {\tt //p-value of the UIT}
\end{algorithmic}
\end{algorithm}


\section{Results}
\label{secResults}
This section describes the reference benchmark dataset that was used
to evaluate the introduction of the flexible independence test selection, 
the results that were obtained, and the tools for the implementation 
of the research.

\subsection{Reference Benchmark Dataset}
The ``ChaLearn cause-effect pair (SUP2)'' is taken for reference as
the recommended benchmark dataset to evaluate the performance
of pairwise causal discovery~\cite{Guyon13}.
It comprises pairs of artificially generated dependent 
variables with different types 
(numerical, categorical and binary) for the full 
causal discovery task (orientation, independence and confounding).
The dataset features a balanced number of unique values across all 
classes and includes around 6000 pairs. In terms of data type balance,
the majority is comprised of numerical and mixed variable pairs. Finally, 
the average median length of an instance is around 2000 values.

\subsection{Performance Scores}
The accuracy classification score, i.e., the total rate of correct
predictions statistically given by the overall amount of
true positives and true negatives,
is used in this research following the previous benchmark
evaluation approaches~\cite{Guyon13}, yielding a baseline around 
65$\pm$2\% for the supervised learning setting~\cite{Giori22}.
To further assess the stability of our scores in the unsupervised
scenario, statistical
bootstrapping is introduced in line with this former work.
Moreover, Gaussianity is asserted with the Lilliefors normality
test~\cite{Lilliefors67},  and the main descriptive
statistics are extracted despite the low amount of available performance
samples, well under 30, which are commonly
required to obtain reliable estimations~\cite{Lejeune10}.
Finally, in terms of statistically significant comparisons, the t-test has been
used~\cite{Student08}.
Table~\ref{tabRes} shows the results of the experiments.
Note that they are presented separately by data type and independence
test, whereas the proposed RESFIT algorithm automatically integrates
them. This is done for illustrative purposes and for enriching 
the ensuing discussion.

\begin{table}[h] \small  
	\begin{center}  
		\caption{Descriptive statistics (mean\,$\pm$\,std)
		for the 
		accuracy classification scores regarding the different
		data types and unconditional independence tests.
		The best results (i.e., the highest value within each
		data type stratum) are shown in boldface.
		After checking Gaussian normality in the performance scores
		(with the exception of TIC for the Categorical data type,
		shown in italics), the
		p-values of the t-test are also provided to assert the
		statistical significance of their difference.\\}
	\label{tabRes}
	\begin{tabular}{ l | c c | c}
		\hline
		\textbf{Data Types} & \textbf{$\chi^2$} & \textbf{TIC} & \textbf{p-value}\\ 
		\hline
		Categorical & 0.2867 $\pm$ 0.1612 & \textbf{\textit{ 0.3500 $\pm$ 0.1329}} & 0.0000\\
		Binary & 0.5027 $\pm$ 0.1854 & {\bf 0.5336 $\pm$ 0.1375} & 0.0014\\
		Numerical & {\bf 0.4300 $\pm$ 0.0260} & 0.3683 $\pm$ 0.0211 & 0.0000\\
		Mixed & {\bf 0.3215 $\pm$ 0.0392} & 0.2456 $\pm$ 0.0219 & 0.0000\\ \hline
		Total & 0.3850 $\pm$ 0.0181 & 0.3211 $\pm$ 0.0136 & 0.0000\\ \hline
	\end{tabular}
	\end{center}
\end{table}

The first straightforward conclusion that the results show is that,
in all cases, there is a statistically significant difference between
the accuracy averages for the two unconditional independence tests
within each data type stratum. Therefore, introducing a selection 
action on the test function is expected to impact 
on the discovery of causal associations.
Additionally, TIC yields a slightly smaller variance in all
the scores. Finally, when numerical data types are present,
the standard deviation in accuracy drops an order of magnitude.

The code for this research is available 
here\footnote{Code repository: https://github.com/atrilla/ccia24}.
The next section discusses the global results with respect to the
perspective of a causal effect, and addresses the limitations
of the proposed approach.

\section{Discussion}
\label{secDisc}
This section delves into the finer details of the results that
were obtained, and sheds light on the actual value introduced by
the flexibility in selecting the unconditional independence test
for causal discovery.

\subsection{Average Causal Effect of the Flexible Test Selection}
To properly quantify the impact of introducing the flexibility
on the unconditional independence test, this section 
treats this selection action as a ``treatment'' variable $X$ 
and studies its impact on the ``outcome'' discovery accuracy
score $Y$.
The Average Causal Effect (ACE) of $X \rightarrow Y$ is
expressed using the following counterfactual notation 

\begin{equation}
\begin{aligned}
ACE = & \mathbb{E} [Y_1 - Y_0] = \mathbb{E} [Y_1] - \mathbb{E} [Y_0] = \mathbb{E} [Y(X=1)] - \mathbb{E} [Y(X=0)] \\
	= & \mathbb{E} [Y|X=1] - \mathbb{E} [Y|X=0] \ ,
\end{aligned}
\label{eqACE}
\end{equation}

where $Y_x$ refers to the value the $Y$ accuracy result would have
if $X$ was set to $x$, i.e., $Y(X=x)$. Here, $X$
could take the values $x=1$ to indicate
flexibility of independence test, and $x=0$ to indicate no
preference (i.e., the null random choice). The conditionals that
eventually follow are the values that are actually observed in the results.
Note that all these expected quantities can be exactly calculated 
because the different experiments can be conducted on the same data
(also dispelling any doubts about latent confounding).
This setting is thus not subject to the fundamental problem of 
causal inference where only one of the potential outcomes
can be observed~\cite{Holland86}.

Equation~(\ref{eqACE}) is shown to be an unbiased estimator for the
ACE~\cite{Neyman23}. For computing the expectations in the
causal effect comparison, a weighted mixture of Gaussian random
variables is required.
The quantity of the reference potential outcome 
$\mathbb{E} [Y_0]$, where no smart selection occurs, is given by
the arithmetic mean value of the total results, and it yields
an accuracy score of 0.3531$\pm$0.0357. Alternatively, the potential outcome
$\mathbb{E} [Y_1]$, where the smart flexible test selection is introduced, 
is given by the average of the best performance results among 
the different variable
types, i.e., a sum weighted by their balance in the dataset,
and it yields a value of 0.3866$\pm$0.0690. Therefore, the ACE
is of 3.36\%, and this estimated difference is statistically significant.
However, the absolute results are far from the 65\% baseline, which
suggests that the original supervised approach may have leaked
statistical patterns beyond causation. Although the unsupervised
learning approach does not seem to offer any pragmatic quantifiable
advantage, we believe that its adoption is a must for tackling the
inherently hard exploratory objective of causal discovery and avoid
any qualms with regards to learned biases.

\subsection{Limitations}
The realization that TIC, which was conceived for
numerical data, performed better on discrete data is surprising.
Also, $\chi^2$, which is defined for discrete data, performed 
better on numerical data. This counterintuitive behavior 
where theory and practice disagree may
suggest that the independence tests that were used are subject 
to some inherent limitations such as the sample size. 
We regard the shortage of samples to be the main limitation
of our approach on real-world data applications such as the
Tuebingen dataset~\cite{Mooij16}.

Finally, while the approach that we propose is able to welcome
any testing technique, we selected RESIT flexibly paired with TIC and 
$\chi^2$ following the advice of Peters, Reshef, and 
colleagues~\cite{Peters17,Reshef16}. To the best
of our knowledge, this limited choice should be sufficient to 
generally cover the spectrum of approaches.

\section{Conclusion}
\label{secConc}

Environments with heterogeneous data pose challenging questions to
the discovery of causal structure, and leveraging one single method may lead to erroneous results in general.
In this work, we propose an
unsupervised pairwise approach using linear functional causal 
models and different unconditional independence tests based on 
Mutual Information measures, the selection of which 
is driven by the nature of data and the empirical results 
from a reference benchmark.
The introduction of this independence test selection flexibility is
estimated to have a positive and statistically significant average
causal effect over 3\% in accuracy.
These scores may establish the first 
standard baseline for this kind of flexible focus, but more
research is needed because the limitations of the
data size and the statistical techniques can result in 
counterintuitive results.

\bibliographystyle{vancouver}
\bibliography{ccia24.bib}

\end{document}